\documentclass[runningheads]{llncs}

\usepackage{eccv}

\usepackage{eccvabbrv}

\usepackage{graphicx}
\usepackage{booktabs}

\usepackage[accsupp]{axessibility}  
\usepackage{booktabs}
\usepackage{multirow}
\usepackage{makecell}

\usepackage{hyperref}

\usepackage{orcidlink}
\usepackage{float}
\usepackage{caption}
\begin{document}

\title{Pretext Matters: An Empirical Study of SSL Methods in Medical Imaging} 

\titlerunning{Pretext Matters: SSL in Medical Imaging}

\author{Vedrana Ivezić\textsuperscript{*}\inst{1,2}\orcidlink{0000-0001-5986-8929} \and
Mara Pleasure\textsuperscript{*}\inst{1,2}\orcidlink{0000-0003-0496-5806} \and
Ashwath Radhachandran\inst{1,6}\orcidlink{0000-0001-8793-0312} \and
Saarang Panchavati\inst{1}\orcidlink{0000-0003-1658-0364} \and
Shreeram Athreya\inst{1,2,7}\orcidlink{0000-0001-5051-2723} \and
Vivek Sant\inst{3}\orcidlink{0000-0002-7688-7790} \and
Benjamin Emert\inst{4} \and
Gregory Fishbein\inst{4} \and
Corey Arnold\inst{1,2,4,5,6}\orcidlink{0000-0002-4119-8143} \and
William Speier\inst{1,2,6}\orcidlink{0000-0002-0890-8684}
}

\authorrunning{V.~Ivezić et al.}

\institute{Biomedical AI Research Lab, University of California, Los Angeles, Los Angeles, CA, USA \and
Department of Radiology, University of California, Los Angeles, Los Angeles, CA, USA \and
Division of Endocrine Surgery, University of Texas Southwestern Medical Center, Dallas, TX, USA \and
Department of Pathology, University of California, Los Angeles, Los Angeles, CA, USA \and
Department of Computational Medicine, University of California, Los Angeles, Los Angeles, CA, USA \and
Department of Bioengineering, University of California, Los Angeles, Los Angeles, CA, USA \and
Department of Electrical and Computer Engineering, University of California, Los Angeles, Los Angeles, CA, USA 
\\[0.5em]
\textsuperscript{*}equal contribution
}


\maketitle

\begin{abstract}
Though self-supervised learning (SSL) has demonstrated incredible ability to learn robust representations from unlabeled data, the choice of optimal SSL strategy can lead to vastly different performance outcomes in specialized domains. Joint embedding architectures (JEAs) and joint embedding predictive architectures (JEPAs) have shown robustness to noise and strong semantic feature learning compared to pixel reconstruction-based SSL methods, leading to widespread adoption in medical imaging. However, no prior work has systematically investigated which SSL objective is better aligned with the spatial organization of clinically relevant signal. In this work, we empirically investigate how the choice of SSL method impacts the learned representations in medical imaging. We select two representative imaging modalities characterized by unique noise profiles: ultrasound and histopathology. When informative signal is spatially localized, as in histopathology, JEAs are more effective due to their view-invariance objective. In contrast, when diagnostically relevant information is globally structured, such as the macroscopic anatomy present in liver ultrasounds, JEPAs are optimal. These differences are especially evident in the clinical relevance of the learned features, as independently validated by board-certified radiologists and pathologists. Together, our results provide a framework for matching SSL objectives to the structural and noise properties of medical imaging modalities.
  
\end{abstract}

\section{Introduction}
\label{sec:intro}

Self-supervised learning (SSL) is the dominant paradigm for training foundation models on large-scale unlabeled image datasets~\cite{jing_self-supervised_2019}. SSL methods are distinguished by their pretext tasks: pixel reconstruction~\cite{mae, tong_videomae_2022}, view-invariance through augmentation~\cite{dinov1,dinov2,dinov3}, and latent prediction~\cite{ijepa}.  Image encoders trained with these pretext tasks have demonstrated strong performance in learning representations that are robust to distribution shift and generalizable to unseen data \cite{dinov3, ijepa, echojepa, ultradino, chen2024towards, vorontsov2024foundation, dosovitskiy_image_2021}.

Recent work in natural imaging has explored under which data conditions each pretext task is most effective. Pixel-level reconstruction excels when signal-to-noise ratio is high; when noise dominates, pixel reconstruction methods model observed variance rather than underlying semantics, yielding less useful representations~\cite{balestriero_learning_2024, assel_joint_2025}.  Joint-embedding architectures (JEAs) address this by predicting in latent space, where irrelevant features are discarded rather than reconstructed~\cite{balestriero_learning_2024, ijepa}. However, JEAs depend on hand-crafted augmentations to define invariances, making them optimal when effective augmentations are identifiable a priori~\cite{dinov1, assel_joint_2025}. Joint-embedding predictive architectures (JEPA) are similar to reconstruction based models but reconstruct targets in the latent space and consequently, like JEAs, are more robust to noise \cite{ijepa}. JEPAs do not use hand-crafted data augmentations and as a result have less inductive bias, resulting in greater generalization to unseen data \cite{ijepa}.

While the trade-offs among these methods have been studied in natural imaging, there has been limited exploration of the influence of pretext tasks in medical imaging \cite{zeng_self-supervised_2024}. Existing applications typically evaluate downstream performance against narrow baselines \cite{echojepa, ultradino}, rather than systematically investigating whether learned representations capture clinically relevant structures and details over modality-specific noise. In this study, we compare MAE~\cite{mae}, DINOv3~\cite{dinov3}, and I-JEPA~\cite{ijepa} across two imaging modalities with distinct signal structure and noise characteristics: ultrasound and histopathology. We find that the pretext task is critically relevant to the representation space that is learned, directly influencing downstream results and interpretability.

\paragraph{\textbf{Contributions}}
\begin{itemize}
    \item We present the first systematic empirical study comparing modern self-supervised learning objectives for medical imaging, analyzing how the choice of pretext task influences the spatial structure of learned representations.

    \item We show that different SSL objectives encode complementary visual cues: DINOv3 preferentially captures fine-grained local structures, while I-JEPA better preserves global spatial relationships.

    \item We derive practical guidelines for selecting SSL methods in clinical imaging, demonstrating that downstream performance depends critically on aligning the SSL objective with the spatial characteristics of clinically relevant features.
\end{itemize}

\section{Background}
\label{sec:bkg}

\begin{figure}[!htbp]
  \centering
  \includegraphics[trim={0cm 0cm 0cm 0cm}, clip, width=\textwidth]{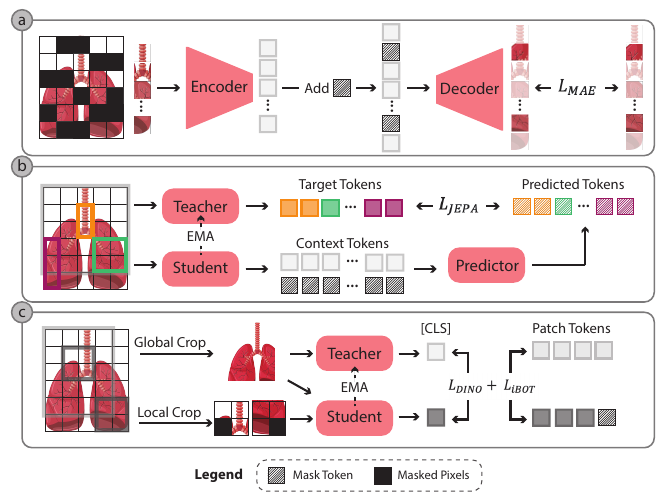}
  \caption{SSL pretext tasks on a lung image for (a) MAE, (b) I-JEPA, and (c) DINOv3.
  }
  \label{fig:pretext_tasks}
\end{figure}

In this section, we motivate the need for a more principled approach to SSL method selection in medical imaging. We review one representative method from each of the three dominant SSL paradigms~(\cref{fig:pretext_tasks}), describing their architectures, pretext tasks, and implications for the learned representations.

\subsection{Reconstruction} 
MAE extends masked language modeling to vision by masking a large proportion of image patches and training an asymmetric encoder-decoder to reconstruct the missing pixels~\cite{mae}. 
The encoder only processes the visible patches (25\%) while a lightweight decoder reconstructs the full image from the encoder's latent output and learnable masked tokens, optimized with mean-squared error loss \cite{mae}. The original implementation demonstrated that pretext tasks that had been successful in self-supervised language learning tasks could be extended to self-supervised image representation learning \cite{mae}. MAE learns localized spatial patterns by inferring missing information from a sparse context; however, optimizing for pixel-level reconstruction often forces the model to waste substantial capacity on high-frequency details which lack high-level semantic value~\cite{ramesh2021zero, ijepa, bao2021beit}. 

\subsection{Joint Embedding Architectures}
The Distillation with No Labels (DINO) family of models learns representations through knowledge distillation between a student and teacher with identical architectures~\cite{dinov1, dinov2, dinov3}. Given an input image, sets of global and local crops are generated. Both networks receive the global crops, but only the student receives local crops. The student is trained to match the teacher's probability distribution via cross-entropy minimization, while the teacher is updated with an exponential moving average (EMA) of the student's weights. By aligning views from the student with the teacher's global views, the model learns relationships between local salient features and the broader semantic context~\cite{dinov1}. DINOv2 \cite{dinov2} incorporated a masked image modeling (MIM) objective from iBOT~\cite{zhou_ibot_2022} to strengthen the local fine-grained features. DINOv3, introduced further training optimizations for large-scale models and incorporated gram anchoring to improve dense feature learning during prolonged training. We adopt the DINOv3 release due to these training optimizations but omit gram anchoring given our small dataset and model scale. For clarity, we refer to DINOv3 as a JEA throughout this paper, as its primary architectural paradigm is view-invariance through self-distillation, while noting that its MIM component introduces a complementary inductive bias toward local features.



\subsection{Joint embedding predictive architectures}

Joint Embedding Predictive Architectures (JEPA) attempt to bridge the conceptual gap between masked modeling and JEA approaches. In image-based JEPA (I-JEPA), instead of reconstructing raw pixels or comparing views, the encoder learns by predicting missing localized information within an abstract representation space \cite{ijepa}. Given a single visible context region, a predictor network is trained to predict the latent representations of masked target blocks, where target embeddings are produced by an EMA-updated teacher encoder \cite{ijepa}. By predicting in latent space rather than pixel space, I-JEPA inherits the noise robustness of JEAs while avoiding dependence on hand-crafted augmentations. Instead of learning view invariance, I-JEPA must model spatial relationships and structural dependencies to predict masked targets, biasing the model towards learning spatial structures \cite{ijepa}.

\subsection{Influence of pretext tasks on learned features}

The choice of pretext task shapes what a model learns to represent. Understanding this relationship is central to our proposed framework for SSL method selection in medical imaging. 

Reconstruction-based approaches such as MAE optimize directly in pixel-space, biasing the model toward features that explain the majority of input variance. When task-irrelevant features such as noise or imaging artifacts carry high magnitude, reconstruction objectives force the model to allocate capacity to variance that is uninformative for downstream tasks~\cite{assel_joint_2025}. Recent approaches shift the pretext task from predicting in the pixel space to predicting contextualized representations \cite{baevski_data2vec_2022, lecun2022path}. By operating in representation space, the encoder is no longer forced to model every low-level detail and can instead capture abstract dependencies while discarding irrelevant variance. Theoretical analysis of deep linear models corroborates this intuition: latent-space prediction models preferentially learn high-influence features, those with large regression coefficients for downstream tasks, while suppressing noisy features during training~\cite{littwin_how_2024}. Consequently, latent space prediction architectures (both JEAs and JEPAs) are theoretically preferable for datasets dominated by high-frequency noise, such as medical imaging~\cite{assel_joint_2025}. 

Even within latent space architectures, the specific pretext task determines what features are captured. JEAs optimize alignment between augmentations, encouraging global semantic understanding and sensitivity to localized patterns. These properties make JEAs highly effective at representing local features with global relevance. However, invariance-based learning may sacrifice sensitivity to continuous macro-structures, such as organ boundaries, whose spatial relationships carry diagnostic meaning. JEPAs, by contrast, must model spatial relationships and structural dependencies to predict masked latent targets from visible context. This property makes JEPAs theoretically better suited to capturing macro-structural information, though they may smooth over fine-grained textures~\cite{ijepa}.

Ultimately, the pretext task is an inductive bias for the learned representations. The ability of an SSL model to capture task-relevant features depends on the alignment between the pretext objective and the spatial organization of clinically meaningful signal in the image. 

\subsection{SSL in medical imaging}
SSL has seen rapid adoption in medical imaging, driven by the availability of large unlabeled datasets and the high cost of expert annotation~\cite{zeng_self-supervised_2024, bundele_evaluating_2025, vorontsov2024foundation, chen2024towards, wang2022transformer, xu2024whole, filiot2024phikon,jiao_usfm_2024, megahed_usf-mae_2025, zhang_fully_2025,ultradino, ellis_self-supervised_2025,echojepa,radhachandran_us-jepa_2026, ivezic_cytofm_2025}. Across domains, different pretext tasks have achieved competitive performance, but it remains unclear how the choice of pretext task affects learned representations (we review domain-specific SSL applications in Supplementary~\cref{sec:sup_ssl_med_imaging}). This unresolved question motivates our study: rather than scaling a single paradigm, we investigate whether pretext task selection should be guided by the relevant structural properties of the target imaging modality and the target task.

\section{Preliminaries}
\label{sec:prelim} 

\subsection{SSL Pre-training}

We pre-train ViT-small models with each SSL paradigm on large-scale histopathology and ultrasound datasets. Downstream evaluations use a frozen version of the model to ensure a fair comparison over the pretext tasks. For each SSL method we use the publicly available implementations and hyper-parameters. More details on training can be found in  Supplementary~\cref{sec:sup_pretrain}.



\begin{table}[H]
\centering
\caption{Overview of downstream tasks. (Acronyms: NAFLD - Non Alcoholic Fatty Liver Disease; LUAD - Lung Adenocarcinoma; LSCC - Lung Squamous Cell Carcinoma; ISUP - International Society of Urological Pathology.) See \ref{sec:sup_datasets} for more info.}
\label{tab:downstream_tasks}
\resizebox{\columnwidth}{!}{
\begin{tabular}{@{}ll|ll@{}}
\toprule
\multicolumn{2}{c}{\textbf{Ultrasound}} & \multicolumn{2}{c}{\textbf{Histopathology}} \\ 
\midrule
\textbf{Dataset} & \textbf{Task} & \textbf{Dataset} & \textbf{Task} \\ 
\midrule

\textbf{MMOTU} & Ovarian Tumor & \textbf{CPTAC-Lung} & LUAD vs LSCC \\
\textbf{AUL} & Liver Mass Malignancy & \textbf{PLCO-Lung} & LUAD vs LSCC \\
\textbf{TN5000} & Thyroid Tumor & \textbf{PLCO-Breast} & Breast Cancer Subtype \\
\textbf{Fatty Liver} & Hepatic Steatosis (NAFLD) & \textbf{Ovarian} & Ovarian Cancer Subtype \\
\textbf{POCUS} & Lung Pathology & \textbf{PANDA} & Prostate Biopsy ISUP Grade \\
\textbf{BUTTERFLY} & Multi-region Anatomical Organ &  &  \\
\textbf{GBCU} & Gallbladder Lesion Malignancy &  &  \\
\textbf{BUSBRA} & Breast Tumor Malignancy &  &  \\

\bottomrule
\end{tabular}}
\end{table}

\subsection{Pre-training Datasets}

For ultrasound, we follow dataset curation as outlined in a recent ultrasound foundation model approach \cite{radhachandran_us-jepa_2026}. The resulting pretraining corpus includes 4.7 million frames, which are extracted from three ultrasound data formats: cine videos, 3D volumes, and static frames. The dataset captures a diverse range of anatomy commonly imaged using ultrasound, including the heart and associated vascular structures, thyroid, liver, and kidney. The pretraining corpus is intentionally sourced from different clinical contexts, ranging from larger academic repositories to smaller curated datasets, ensuring that models are exposed to a range of real-world clinical environments and scanning conditions. 

For pathology SSL models, pretraining was performed on whole-slide images (WSIs) obtained from The Cancer Genome Atlas (TCGA). All cancer types with available WSIs were incorporated into the pretraining dataset, which consists of over 5 million patches extracted from 9,498 WSIs across 28 cancer types. All 256 $\times$ 256 patches were extracted at 20$\times$ magnification with no overlap, resulting in a median of 492 patches per WSI. Proportional sampling was used to create the 5 million patch dataset to incorporate each cancer type relative to its abundance in the TCGA.

\subsection{Downstream Datasets}

Downstream tasks for the models trained on ultrasound are established in line with the evaluation benchmark used in \cite{radhachandran_us-jepa_2026} which expands on a previously published evaluation dataset suite, UltraBench \cite{tupper_revisiting_2025}. The downstream evaluations span a range of clinical endpoints including oncological and non-oncological targets. We assess broader anatomical identification through a nine-class organ differentiation task. See~\cref{tab:downstream_tasks} for specific task breakdown. SSL methods trained on histopathology are evaluated on five public WSI datasets spanning four cancer types: ovarian, breast, lung, and prostate. We evaluated the performance of each model on subtype prediction, using the slide-level or patient-level label (when available) to predict the subtype or grade of the cancer. See \ref{sec:sup_datasets} for additional data details.

\section{Experiments}
\label{sec:experiments} 


\begin{figure}[!tbp]
  \centering
  \includegraphics[trim={0cm 0cm 0cm 0cm}, clip, width=\textwidth]{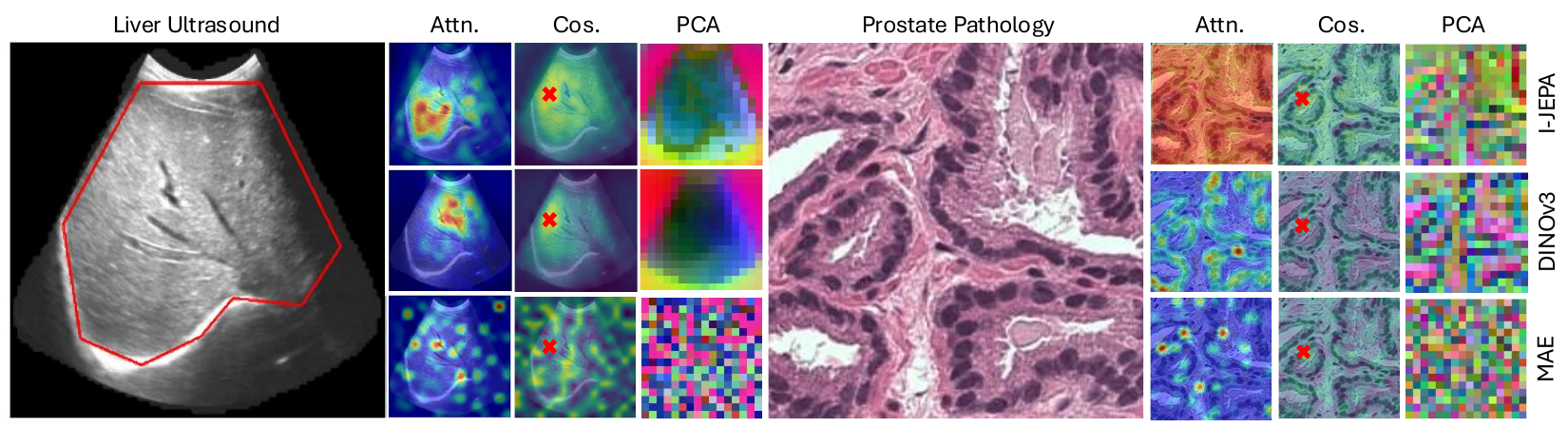}
  \caption{Example visualizations including attention maps, cosine similarity maps (anchor points denoted by red X), and PCA maps for both ultrasound and pathology.
  }
  \label{fig:hero_plot}
\end{figure}

We evaluate three SSL methods for representation learning in ultrasound and histopathology. For ultrasound, we assess learned features using linear probing. For histopathology, due to the size of WSIs, we use attention-based multiple instance learning (ABMIL) to aggregate and classify patches using slide-level or patient-level labels~\cite{ilse2018attention}.

Beyond downstream performance, we visualize the learned representation spaces using three complementary methods~(\cref{fig:hero_plot}). Attention maps reveal which image regions the pretrained model treats as important. Cosine similarity maps identify which patches the model places close to an anchor point in representation space. PCA maps reduce patch embeddings to three components~\cite{dinov3}, visualizing relationships between patches in the learned feature space. In I-JEPA, attention maps are computed as the mean over the attentions from each patch serving as a query. In DINOv3 and MAE, however, the attention maps are computed using the \texttt{[CLS]} token as the query, following their original implementations (mean patch attention is provided in Supplementary~\cref{sec:sup_attn_maps}). Patch level features were used for both cosine similarity and PCA maps for each model.

\subsection{SSL in Ultrasound}

\paragraph{\textbf{Task-specific classification results.}}


~\cref{tab:ultrasound_results} presents linear probing results for eight classification tasks across multiple organs and diseases. For each task, a single linear layer maps frozen SSL embeddings to output classes, trained for up to 300 epochs (early stopping patience = 30) with a 0.001 learning rate.

The DINOv3 pretext task outperforms I-JEPA and MAE on five out of eight tasks while I-JEPA is the best on the three remaining tasks. While DINOv3 outperforms I-JEPA by at most 2.64\% AUROC, on the Fatty Liver task I-JEPA outperforms DINOv3 by more than 10\%. Interestingly, on the region classification task (Butterfly), I-JEPA slightly outperforms DINOv3 but the class-specific performance varies greatly between models for certain ultrasound regions. Both models perform similarly well on identifying ultrasounds of the inferior vena cava, bladder, thyroid, Morison's pouch, and the PLAX view of the heart. I-JEPA outperforms DINOv3 when identifying the carotid and lung ultrasounds by 16.16\% and 15.15\% (F1-score), respectively. DINOv3 is better at identifying both two- and four-chamber heart ultrasounds by 5.93\% and 9.05\%, respectively. PCA visualizations further demonstrate I-JEPA's ability to separate macro structures.


\begin{table}[!htbp]
\centering
\caption{\small Performance comparison across SSL methods on eight ultrasound classification tasks averaged across five seeds. All values are reported as percentages (\%).}
\resizebox{\columnwidth}{!}{%
\begin{tabular}{ccccccc}
 & \multicolumn{2}{c}{\textbf{MAE}} & \multicolumn{2}{c}{\textbf{DINOv3}} & \multicolumn{2}{c}{\textbf{I-JEPA}} \\ \toprule
Dataset & \textit{AUROC (\%)} & \textit{F1 (\%)} & \textit{AUROC (\%)} & \textit{F1 (\%)} & \textit{AUROC (\%)} & \textit{F1 (\%)} \\ \midrule
\multicolumn{1}{c|}{\textbf{MMOTU}} & \multicolumn{1}{c|}{78.43$\pm$0.12} & \multicolumn{1}{c|}{38.04$\pm$1.4} & \multicolumn{1}{c|}{\textbf{84.57$\pm$0.07}} & \multicolumn{1}{c|}{\textbf{50.31$\pm$0.65}} & \multicolumn{1}{c|}{82.59$\pm$0.06} & 42.03$\pm$0.58 \\
\multicolumn{1}{c|}{\textbf{AUL}} & \multicolumn{1}{c|}{81.86$\pm$0.21} & \multicolumn{1}{c|}{62.52$\pm$2.1} & \multicolumn{1}{c|}{\textbf{86.77$\pm$0.23}} & \multicolumn{1}{c|}{\textbf{66.33$\pm$0.52}} & \multicolumn{1}{c|}{86.47$\pm$0.21} & 64.71$\pm$0.81 \\
\multicolumn{1}{c|}{\textbf{TN5000}} & \multicolumn{1}{c|}{69.88$\pm$0.2} & \multicolumn{1}{c|}{58.25$\pm$2.2} & \multicolumn{1}{c|}{72.36$\pm$0.42} & \multicolumn{1}{c|}{63.34$\pm$1.5} & \multicolumn{1}{c|}{\textbf{75.60$\pm$0.12}} & \textbf{65.83$\pm$1.6} \\
\multicolumn{1}{c|}{\textbf{Fatty Liver}} & \multicolumn{1}{c|}{65.89$\pm$0.6} & \multicolumn{1}{c|}{50.76$\pm$5.7} & \multicolumn{1}{c|}{86.66$\pm$1.5} & \multicolumn{1}{c|}{64.06$\pm$1.7} & \multicolumn{1}{c|}{\textbf{98.70$\pm$0.26}} & \textbf{92.80$\pm$1.2} \\
\multicolumn{1}{c|}{\textbf{POCUS}} & \multicolumn{1}{c|}{96.12$\pm$0.19} & \multicolumn{1}{c|}{91.01$\pm$0.11} & \multicolumn{1}{c|}{\textbf{99.49$\pm$0.07}} & \multicolumn{1}{c|}{\textbf{94.00$\pm$0.19}} & \multicolumn{1}{c|}{97.18$\pm$0.06} & 90.48$\pm$0.52 \\
\multicolumn{1}{c|}{\textbf{BUTTERFLY}} & \multicolumn{1}{c|}{98.96$\pm$0.11} & \multicolumn{1}{c|}{88.34$\pm$0.98} & \multicolumn{1}{c|}{99.66$\pm$0.03} & \multicolumn{1}{c|}{90.94$\pm$0.27} & \multicolumn{1}{c|}{\textbf{99.71$\pm$0.01}} & \textbf{93.15$\pm$0.35} \\
\multicolumn{1}{c|}{\textbf{GBCU}} & \multicolumn{1}{c|}{66.23$\pm$0.18} & \multicolumn{1}{c|}{48.48$\pm$1.1} & \multicolumn{1}{c|}{\textbf{80.00$\pm$0.55}} & \multicolumn{1}{c|}{\textbf{64.45$\pm$1.7}} & \multicolumn{1}{c|}{77.80$\pm$0.28} & 59.22$\pm$3.7 \\
\multicolumn{1}{c|}{\textbf{BUSBRA}} & \multicolumn{1}{c|}{76.63$\pm$0.2} & \multicolumn{1}{c|}{67.30$\pm$2.4} & \multicolumn{1}{c|}{\textbf{77.92$\pm$0.36}} & \multicolumn{1}{c|}{\textbf{71.51$\pm$1.0}} & \multicolumn{1}{c|}{76.78$\pm$0.07} & 67.35$\pm$0.24 \\ \midrule
\end{tabular}%
}

\label{tab:ultrasound_results}
\end{table}

\begin{figure}[!tbp]
  \centering
  \includegraphics[trim={0cm 4cm 0cm 0cm}, clip, width=0.9\textwidth]{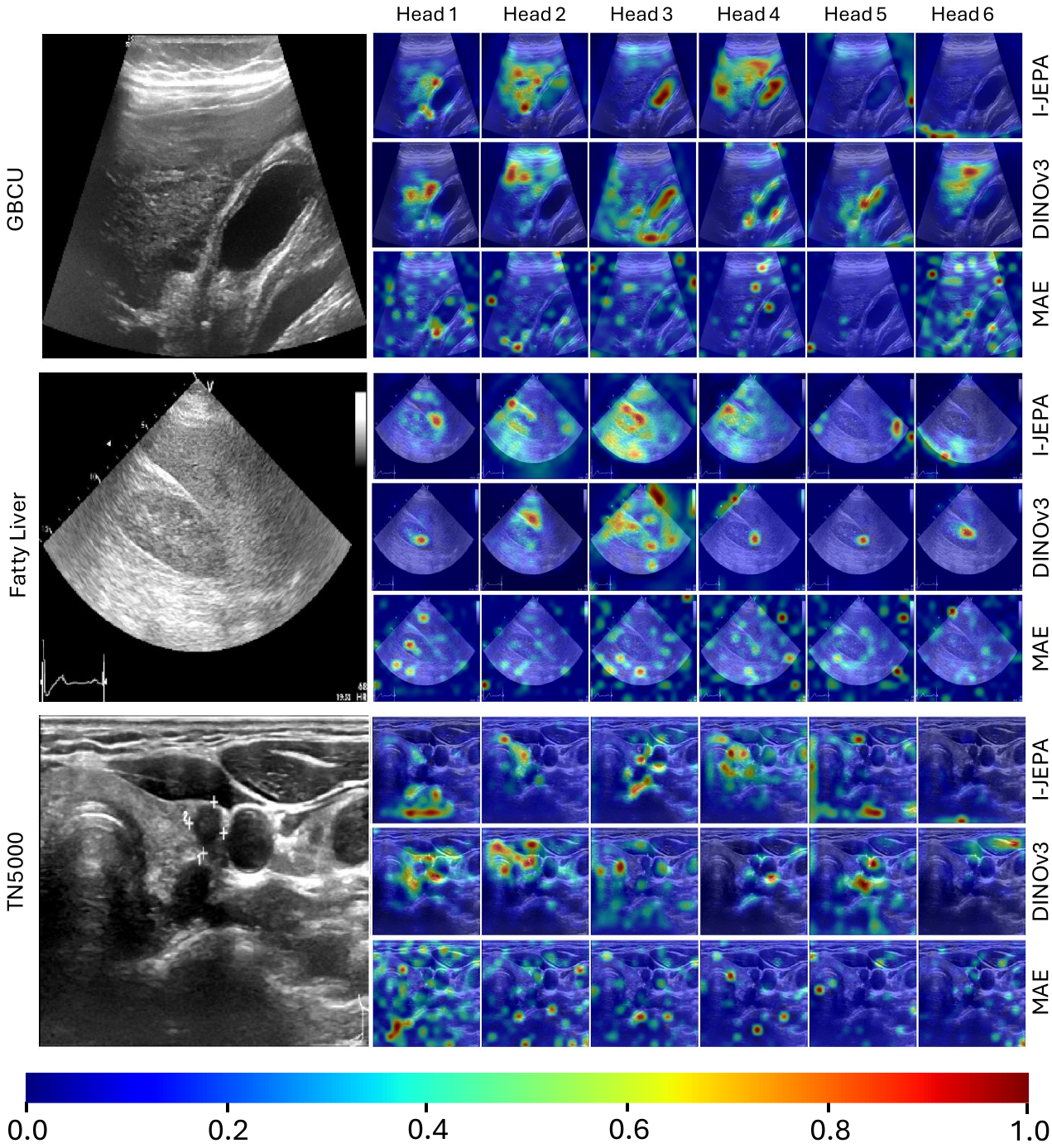}
  \caption{Attention heatmaps from each attention head of the final ViT layer across models pre-trained on ultrasounds. The dark oval area in the GBCU image is the gallbladder with the liver region to the left. The oval area of the Fatty Liver image is the kidney, the white line above is the diaphragm, and the shaded region to the right is the liver. The TN5000 image contains a thyroid nodule marked with calipers.
  }
  \label{fig:attention_ultrasound}
\end{figure}

\paragraph{\textbf{Attention and feature visualizations.}}
A board-certified radiologist observed that DINOv3 and I-JEPA attend to clinically relevant regions, while MAE exhibits diffuse attention with limited clinical relevance (\cref{fig:attention_ultrasound}). DINOv3 focuses on localized structures and high-contrast boundaries, such as thyroid tissue edges and the kidney in the Fatty Liver ultrasound. In contrast, I-JEPA attends more broadly to entire structures, including the liver parenchyma in the GBCU and Fatty Liver ultrasounds.


\begin{figure}[!btp]
  \centering
  \includegraphics[trim={0cm 2cm 0cm 0cm}, clip, width=\textwidth]{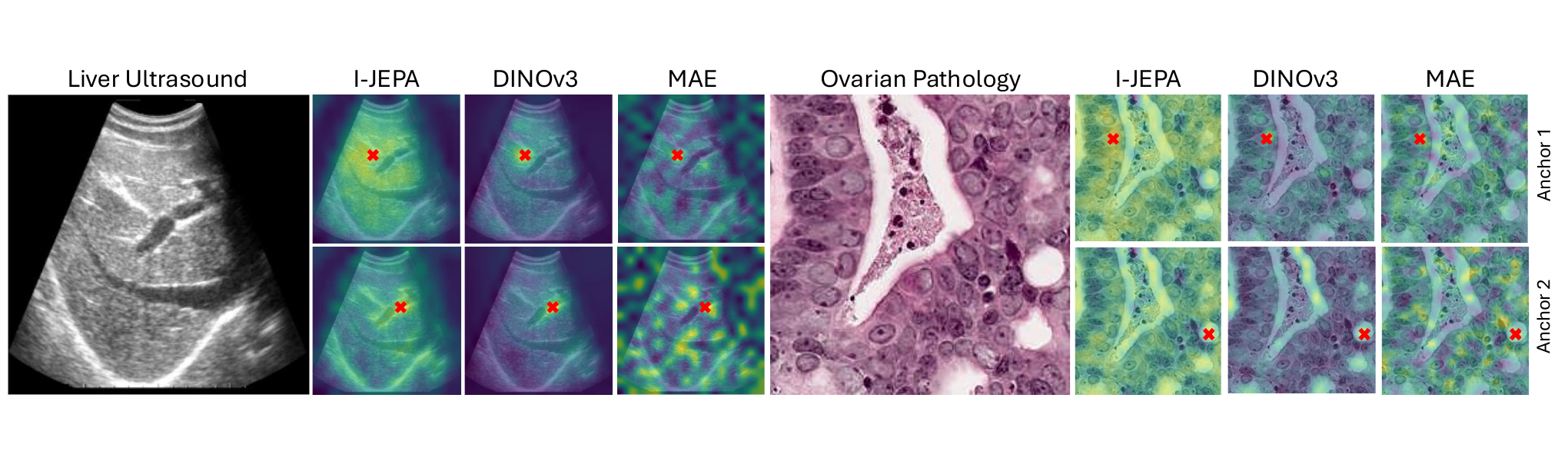}
  \caption{Cosine similarity maps computed in relation to anchor patches (red X). 
  }
  \label{fig:cos_sims}
\end{figure}

\paragraph{\textbf{Anatomical understanding.}}
In addition to attending to the present organs, I-JEPA learns structural similarity within organs, as seen in cosine similarity in the liver~(\cref{fig:cos_sims}). We quantify the ability of I-JEPA compared to DINOv3 to effectively segment the liver in the AUL dataset~(\cref{fig:hero_plot}). For each anchor point selected within the liver region, the area under the receiver operating characteristic (AUROC) is calculated where the segmentation mask serves as the true label and the cosine similarities with respect to the anchor serve as the predicted probabilities. Three different anchors are selected for each image in the dataset and their AUROCs are averaged to get an image level score. The AUROC distributions from I-JEPA and DINOv3 are compared using the Wilcoxon signed-rank test. I-JEPA produces feature clusters that are significantly more semantically consistent and that correspond to anatomical boundaries ($p = 5.76 \times 10^{-21}$). We also compare cosine similarity maps across anchors within an image to test whether the models identify similar regions within the organ. Correlations are computed pairwise across five selected anchors within the liver and averaged. I-JEPA is significantly better at learning the semantic coherence of anatomical structures than DINOv3 ($p = 7.45 \times 10^{-26}$ by Wilcoxon signed-rank test).


\begin{table}[H]
\centering
\caption{\small Performance comparison across SSL methods on five histopathology classification tasks. All values are reported as percentages (\%).}
\resizebox{\columnwidth}{!}{%
\begin{tabular}{ccccccc}
 &
  \multicolumn{2}{c}{\textbf{MAE}} &
  \multicolumn{2}{c}{\textbf{DINOv3}} &
  \multicolumn{2}{c}{\textbf{I-JEPA}} \\ \toprule
Dataset &
  \textit{AUROC (\%)} &
  \textit{F1 (\%)} &
  \textit{AUROC (\%)} &
  \textit{F1 (\%)} &
  \textit{AUROC (\%)} &
  \textit{F1 (\%)} \\ \midrule
\multicolumn{1}{c|}{\textbf{CPTAC-Lung}} &
  \multicolumn{1}{c|}{77.1$\pm$4.2} &
  \multicolumn{1}{c|}{27.2$\pm$33.3} &
  \multicolumn{1}{c|}{\textbf{91.4$\pm$1.0}} &
  \multicolumn{1}{c|}{\textbf{83.3$\pm$1.4}} &
  \multicolumn{1}{c|}{54.8$\pm$1.3} &
  31.6$\pm$29.4 \\
\multicolumn{1}{c|}{\textbf{PLCO-Lung}} &
  \multicolumn{1}{c|}{63.8$\pm$0.9} &
  \multicolumn{1}{c|}{30.6$\pm$13.7} &
  \multicolumn{1}{c|}{\textbf{82.9$\pm$1.3}} &
  \multicolumn{1}{c|}{\textbf{66.8$\pm$5.3}} &
  \multicolumn{1}{c|}{57.4$\pm$8.4} &
  21.8$\pm$26.7 \\
\multicolumn{1}{c|}{\textbf{PLCO-Breast}} &
  \multicolumn{1}{c|}{49.2$\pm$2.2} &
  \multicolumn{1}{c|}{26.5$\pm$6.2} &
  \multicolumn{1}{c|}{\textbf{65.2$\pm$1.1}} &
  \multicolumn{1}{c|}{\textbf{38.7$\pm$8.5}} &
  \multicolumn{1}{c|}{52.3$\pm$6.2} &
  11.6$\pm$8.6 \\
\multicolumn{1}{c|}{\textbf{Ovarian}} &
  \multicolumn{1}{c|}{73.5$\pm$0.7} &
  \multicolumn{1}{c|}{34.1$\pm$7.3} &
  \multicolumn{1}{c|}{\textbf{92.0$\pm$1.8}} &
  \multicolumn{1}{c|}{\textbf{73.3$\pm$2.4}} &
  \multicolumn{1}{c|}{55.2$\pm$4.3} &
  4.5$\pm$2.0 \\
\multicolumn{1}{c|}{\textbf{PANDA}} &
  \multicolumn{1}{c|}{81.9$\pm$0.2} &
  \multicolumn{1}{c|}{44.2$\pm$0.3} &
  \multicolumn{1}{c|}{\textbf{87.0$\pm$0.3}} &
  \multicolumn{1}{c|}{\textbf{53.9$\pm$0.8}} &
  \multicolumn{1}{c|}{64.0$\pm$0.6} &
  22.7$\pm$1.5
\\ \midrule
\end{tabular}%
}
\label{tab:pathology_results}
\end{table}

\subsection{SSL in Pathology}
\cref{tab:pathology_results} presents the ABMIL results for five downstream histopathology tasks. For each dataset, we fine-tuned ABMIL for 20 epochs and selected the best epoch based on validation AUROC. Test performance was evaluated using AUROC and F1 scores. DINOv3 has the highest AUROC and F1 scores across all downstream tasks, consistent with our hypothesis that its local-global alignment objective is well-suited to histopathology. MAE obtained competitive AUROC on the PANDA dataset (81.9 vs.\ DINOv3's 87.0) but did not match DINOv3 on any other task. I-JEPA struggled across all downstream tasks, with uniformly low F1 scores (all $\leq$ 31.6) indicating difficulty distinguishing between classes. 

\begin{figure}[!tbp]
  \centering
  \includegraphics[trim={0cm 0cm 0cm 0cm}, clip, width=0.9\textwidth]{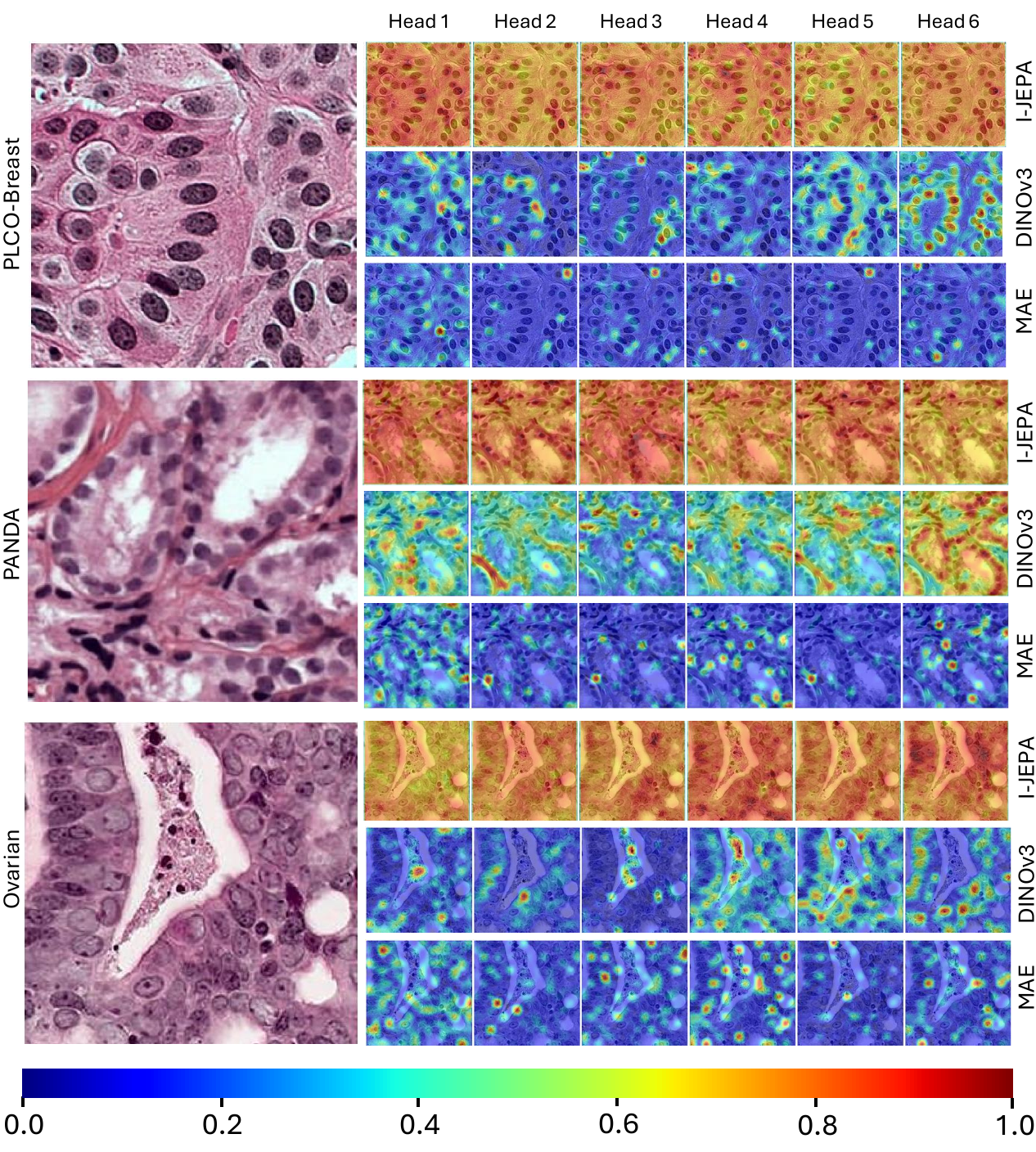}
  \caption{Attention heatmaps from each attention head of the final ViT layer across models pre-trained on histopathology data for three downstream datasets.}
  \label{fig:attention_pathology}
\end{figure}

\paragraph{\textbf{Attention and Feature Similarity Visualizations.}}
To move beyond aggregate performance metrics and assess whether learned representations capture meaningful structure, we visualize patch-level features using multi-modal attention maps, PCA projections, and cosine similarity with selected query points. For each ABMIL model, we extract the top 10 highest-scoring patches per WSI and analyze the corresponding token-level and \texttt{CLS}-token representations~\cref{fig:attention_pathology}. In a blinded study, two pathologists independently reviewed selected visualizations to assess each model's outputs across multiple tissue types.

Both pathologists consistently identified DINOv3 as producing the most interpretable and biologically structured representations. Additionally, both independently highlighted the attention heads as targeting specific biological structures, for example, one head consistently attends to nuclei and basal cells while others preferentially capture luminal cells or regions of dense chromatin. These patterns are consistent across tissue types and across samples within each type. The PCA visualizations reinforce these findings, revealing spatial separation of cellular regions, stromal tissue, and empty glass, with individual glands and appropriate layering being delineated in some plots. These results suggest that DINOv3's pretext task encourages individual heads to specialize in distinct, recognizable features.

I-JEPA's visualizations were characterized by both pathologists as largely uninterpretable in the histopathology setting, with attention maps that are diffuse and non-specific, attending to broad regions rather than discrete structures. Occasional instances of meaningful focus were noted, such as a head attending to small condensed nuclei consistent with lymphocytes, but these trends do not generalize across tissue types. Both pathologists described I-JEPA's representations as capturing coarse structural contrasts such as tissue versus empty space and variations in collagen density, rather than fine-grained cellular features. This pattern suggests that I-JEPA's inability to represent localized cellular morphology likely leads its poor downstream performance. MAE performed worse in the clinical assessment, with both pathologists describing its attention maps and PCA visualizations as random and noisy, consistent with pixel-space objectives that allocate model capacity to staining variation and imaging noise rather than diagnostically relevant morphology.

\section{Discussion}
\label{sec:discussion}
In this work, we investigated whether the choice of self-supervised pretext task leads to differences in learned representations for medical imaging, and whether understanding these differences can guide practitioners in selecting an appropriate method for their medical imaging modality and task. We evaluated three SSL approaches — MAE, DINOv3, and I-JEPA — on two imaging modalities with distinct characteristics: histopathology, where local fine-grained cellular morphology drives diagnosis, and ultrasound, where diagnosis depends on anatomical context, spatial structure, and texture. Our results demonstrate that there is no ideal pretext across all image types; rather, the best method depends on the spatial organization of clinically relevant information within the imaging modality and the specific downstream task.

\paragraph{\textbf{JEPAs and JEAs both capture meaningful clinical signal, but encode different spatial scales.}}

I-JEPA's objective, predicting masked latent regions from visible context, encourages representations that capture spatial relationships and structural continuity, at the cost of fine-grained local features ~\cite{ijepa}. DINOv3's view-invariance objective instead emphasizes discriminative, localized features that distinguish objects from background, and its MIM component adds texture sensitivity~\cite{zhou_ibot_2022} but does not fully recover macro-structural understanding~\cite{dinov2}. MAE, which reconstructs in pixel space, achieves lower performance than both approaches in ultrasound and DINOv3 in pathology. MAE's attention spreads uniformly across patches rather than selectively aggregating task-relevant information, limiting its utility for high-level perception~\cite{przewikezlikowski2025beyond, assel_joint_2025}. These inductive biases manifest clearly in our experiments, with I-JEPA excelling at representing continuous anatomical structures and DINOv3 demonstrating superior fine-grained morphological discrimination.

\paragraph{\textbf{In ultrasound, I-JEPA learns clinically meaningful anatomical structure.}} To make an effective diagnosis using ultrasound, clinicians require both global context to maintain spatial orientation and granular local features to understand pathology. The relationships between anatomical structures provide critical diagnostic information. We observe that I-JEPA captures this larger anatomical context while DINOv3 does not.

Our cosine similarity analysis demonstrates that I-JEPA's representations cluster contiguous anatomical regions like the liver, significantly outperforming DINOv3 at identifying semantic coherence of anatomical structures ($p = 7.45 \times 10^{-26}$). This difference is a result of their respective pretext tasks. While DINOv3's view-invariance objective prioritizes salient features, like contours and contrast, to discriminate objects from background \cite{dinov1, barsellotti2025talking, dinov3}, it struggles with expansive structures like liver parenchyma where discriminative features are less present. Here, DINOv3 attends to the high-contrast boundaries of the diaphragm and kidney rather than the low-contrast liver. In contrast, I-JEPA's predictive objective emphasizes structural consistency. As a result, even without fine-tuning, I-JEPA's attention is diffuse across task-relevant organ regions, capturing both the liver and kidney in the fatty liver images.

This representational difference is reflected in downstream performance on tasks requiring whole-organ understanding. On the fatty liver task where the entire texture of the parenchyma is clinically relevant, I-JEPA substantially outperforms DINOv3 (AUROC 98.70 vs 86.66). In region classification, I-JEPA outperforms DINOv3 in identifying ultrasounds of the carotid arteries and the lungs where organ continuity is a primary differentiating feature rather than the presence of localized objects. While both methods perform comparably on tasks reliant on localized features, I-JEPA's anatomically coherent representations offer a valuable upside when interpretability and alignment with clinical understanding are required.


\paragraph{\textbf{In pathology, DINOv3 learns clinically meaningful fine grained local features.}}

In histopathology, pathologists evaluate cellular features, such as nuclear morphology and chromatin texture, as well as glandular architecture. We observed that DINOv3 captures these cellular features while I-JEPA does not, a finding supported by both quantitative performance and clinical validation. DINOv3 outperformed both alternatives across all five downstream tasks, while I-JEPA's uniformly low F1 scores indicate a fundamental inability to distinguish between classes. Our cosine similarity analysis reveals underlying representational differences; DINOv3 recognizes similar cell types while preserving each cell's unique morphological characteristics, whereas I-JEPA attends primarily to coarse structural contrasts: tissue versus empty space and collagen density. Two pathologists confirmed these findings in a blinded review, consistently identifying DINOv3 as producing biologically interpretable representations. In DINOv3, individual attention heads reliably isolated stroma from epithelium, distinguished nuclear morphologies such as vesicular versus condensed chromatin, and delineated glandular architecture. I-JEPA's attention was characterized as diffuse and non-specific, and did not generalize across tissue types. These findings suggest that in imaging modalities where spatially localized structures and textural features carry diagnostic meaning, JEA-style objectives are preferable. DINOv3's combination of view invariance and masked modeling encourages individual attention heads to specialize in discriminative local features. 

\paragraph{\textbf{How to pick a pretext task?}}

Our findings suggest a framework for selecting SSL methods in medical imaging based on the alignment between the pretext objective and the spatial organization of clinically relevant signal. While our results reflect the combined effect of each method's architecture, augmentation strategy, and objective, the dominant factor distinguishing them is the pretext task, and we observe consistent alignment between the task's inductive bias and the spatial structure of learned features.

Before choosing a pretext task, we should first characterize the spatial organization of the clinically relevant signal within the target modality and its alignment to the downstream task. When diagnostic information is encoded in macro-structures and the spatial configuration, as in ultrasound for fatty liver diagnosis, JEPA-style prediction objectives are preferable. In contrast, when diagnostic information is localized in fine-grained textures, localized features, or morphological variation, as in histopathology, JEAs with masked image modeling are more effective. In both cases, pixel-space reconstruction objectives are suboptimal for medical imaging due to the noise and artifacts typical of clinical data. For modalities where both local and global features carry diagnostic weight, the optimal strategy remains an open question.


\paragraph{\textbf{Limitations and future work.}}

Our evaluation serves as a foundation to understand trade-offs between SSL frameworks in medical imaging. We chose a representative model from each framework but encourage further exploration by the community to leverage this framework to evaluate additional pretext tasks in medical applications. Due to computational constraints we only trained ViT-small models. Future work will focus on understanding how these observations scale with larger models. The methods we chose were evaluated on two distinct imaging modalities but we encourage exploration of these tasks in additional modalities where similar trade-offs should be identified. For instance, this framework should be applied in imaging modalities like CT and MRI where structural understanding especially across image slices is relevant and in cytology where global structure is even less important than in histopathology. Evaluating histopathology at lower magnifications (e.g. 10$\times$) where glandular architecture becomes more prominent relative to individual cellular features, could reveal whether the optimal pretext task shifts as the spatial scale of relevant information changes. Our results underscore the value of cross-domain collaboration with clinicians; expert validation was essential for assessing whether learned representations capture clinically meaningful signal.


\section{Conclusion}
\label{sec:conclusion}

We evaluate three SSL methods in medical imaging and demonstrate that the choice of pretext task directly shapes clinically relevant learned representations. Our analysis provides a framework for matching SSL objectives to the structural properties of medical imaging modalities. We urge practitioners, in collaboration with clinicians, to characterize the diagnostically relevant structure of their images to inform SSL method selection.

\bibliographystyle{splncs04}
\bibliography{main}

\clearpage
\appendix
\label{sec:supplementary}

\section{SSL in Medical Imaging}\label{sec:sup_ssl_med_imaging}

In histopathology, the digitization of whole-slide images (WSIs) resulting in gigapixel images captured at 20$\times$ or 40$\times$ magnification has created a natural application of SSL. Since WSIs are too large for direct model training and pixel-level annotations are expensive, the standard workflow decomposes each slide into sub-patches, compresses each patch into a low-dimensional embedding using a pretrained encoder, and processes the resulting bag of embeddings under a multiple-instance learning (MIL) framework \cite{ilse2018attention, campanella2019clinical}. This pipeline has driven rapid development of histopathology foundation models, where large collections of unannotated WSIs are used for SSL pretraining \cite{vorontsov2024foundation, chen2024towards, wang2022transformer, xu2024whole, filiot2024phikon}.

Early models such as CTransPath (2021) employed a customized Momentum Contrastive learning approach trained on approximately 16 million patches from 32k slides \cite{wang2022transformer}. Since then, the dominant paradigm has converged on DINOv2 pretraining with ViT backbones at increasing scale: UNI and UNI2 from Chen et al. train a ViT-Large on 100k+ diagnostic WSIs \cite{chen2024towards}. Phikon-v1 used iBOT to pretrain their ViT-Base model with Phikon-v2 updating to DINOv2 pretraining as the pretext task \cite{filiot2024phikon}. Virchow from Paige AI trained a ViT-Huge on 2 billion tiles while Prov-GigaPath from Microsoft trained a ViT-Giant on 171k WSIs \cite{xu2024whole, vorontsov2024foundation}. While these efforts have consistently scaled up model size and dataset volume under the DINOv2 framework, comparatively little attention has been paid to understanding what these models actually learn from the joint embedding architecture (JEA) pretext task compared to alternative pretext tasks.

In ultrasound, SSL adoption has been more limited but is rapidly evolving, with a growing body of work applying reconstruction, JEA, and joint embedding predictive architecture (JEPA) frameworks. Initial efforts, such as USFM \cite{jiao_usfm_2024}, USF-MAE \cite{megahed_usf-mae_2025}, and EchoCare \cite{zhang_fully_2025}, use reconstruction-based methods like MAE to learn localized spatial patterns. However, as previously discussed, optimizing for pixel-level reconstruction in low signal-to-noise modalities can be problematic; in ultrasound, the prevalence of speckle noise and acoustic artifacts motivates the need for approaches more robust to this variance. To this end, UltraDINO \cite{ultradino} successfully applies the DINOv2 framework to fetal ultrasound, demonstrating that domain-specific JEA pretraining can outperform larger, general-purpose models. In parallel, others have pursued latent space prediction, with Ellis et al. introducing JEPAs to the domain by adopting V-JEPA for ultrasound video to prioritize temporal information over noisy pixels \cite{ellis_self-supervised_2025}. This approach was followed by EchoJEPA \cite{echojepa}, which demonstrated that JEPAs can develop a functional understanding of cardiac cycles in echocardiography. Most recently, US-JEPA \cite{radhachandran_us-jepa_2026} showed that a JEPA-based objective during pretraining could outperform previous ultrasound foundation models under linear probing.

\section{SSL Model Pre-training}\label{sec:sup_pretrain}

\subsection{Hyperparameters} \label{sec:sup_hyperparam}
The default configuration of hyperparameters were used for each SSL model, except for those listed in~\cref{tab:sup_hyperparameters}. 

The checkpoint used for downstream evaluation for each SSL model was selected when the loss converged. For the ultrasound models, convergence occurred at epoch 20 for MAE, iteration 125000 for DINOv3, and epoch 3 for I-JEPA. In the pathology models, convergence occurred at epoch 20 for MAE, iteration 125000 for DINOv3, and epoch 11 for I-JEPA.

\begin{table}[ht]
    \centering
    \caption{Comparison of Hyperparameters across three Models. Learning rate peak for DINOv3 histopathology model was $1\times10^{-5}$ otherwise all parameters were shared across ultrasound and histopathology.}
    \label{tab:sup_hyperparameters}
    \setlength{\tabcolsep}{10pt}
    \begin{tabular}{@{}lll@{}} \toprule
        \textbf{Model} & \textbf{Hyperparameter} & \textbf{Value} \\ \midrule
        MAE & ViT & ViT-small  \\
            & Batch size & 256 \\ 
                      & Learning rate & $1.5\times10^{-4}$\\
                      & Warmup epochs & 10 \\ \midrule
        DINOv3 & ViT & ViT-small \\
                        & Batch size & 128 \\
                          & Learning rate peak & $1\times10^{-3}$| $1\times10^{-5}$ \\
                          & Learning rate end & $1\times10^{-6}$\\ 
                          & Student warmup epochs & 10 \\ 
                          & Teacher warmup epochs & 10 \\ \midrule
        I-JEPA & ViT & ViT-small  \\ 
               & Batch size & 128 \\ 
               & Predictor depth & 8 \\
               & Predictor heads & 8 \\ 
               & Predictor embedding dimension & 128 \\
               & Warmup & 40 \\ 
                & Learning rate start & 0.0002 \\ 
               & Learning rate & 0.001 \\ \bottomrule
    \end{tabular}
\end{table}



\section{Data Preprocessing}

Pretraining ultrasound images were first converted to grayscale, with minor colored annotations removed via inpainting. To resolve brightness inconsistencies across different scanners, we rescaled pixel intensities using 2nd/98th percentile mapping. Mean and standard deviation were computed over the pretraining dataset and used to standardize each image. Ultrasound images for downstream tasks were already processed and we standardized them using the mean and standard deviation from the pretraining dataset.

All pathology datasets were patched at 224 $\times$ 224 pixels at 20$\times$ magnification and resized to 256 for inference. Images were extracted and stain-normalized using DeepPath~\cite{coudray2018classification} across all datasets. Data was split into train/validation/test partitions of 70/10/20 at the patient level or slide level. Due to computational constraints, we sub-sampled each downstream dataset to 2 million patches and used the same subset for each SSL method. Following previous SSL training in histopathology, we used ImageNet mean and standard deviation to perform further standardization during pretraining.

\section{Downstream Evaluations}

\subsection{Datasets}\label{sec:sup_datasets}

Ultrasound downstream tasks were established in line with the evaluation benchmark used in \cite{radhachandran_us-jepa_2026}. This benchmark expands on a previously published evaluation dataset suite, UltraBench \cite{tupper_revisiting_2025}, with two additional tasks (breast lesion and thyroid nodule malignancy classification) for further anatomical diversity. The other downstream evaluations span a range of clinical endpoints. We evaluated performance on three more oncological targets, with gallbladder, liver and ovarian malignancy classification, as well as non-oncological applications such as fatty liver disease and pulmonary pathology (pneumonia and COVID-19) detection. Finally, we assessed anatomical understanding through a 9-class organ differentiation task.

For the two lung cancer classification tasks we used Hematoxylin \& eosin (H\&E) WSIs obtained from CPTAC and PLCO. Both datasets were set up for binary classification of Lung Adenocarcinoma (LUAD) vs. Lung Squamous Cell Carcinoma (LSCC). For breast cancer subtyping, PLCO-Breast WSIs were also obtained from the PLCO database. Each case was labeled with a subtype of 'lobular', ductal', or 'other' for multiclass classification. Ovarian biopsy samples were obtained from the UBC ovarian subtype classification and outlier detection competition. Each sample was labeled with an ovarian cancer subtype, resulting in 5 classes for multiclass classification. PANDA Core needle biopsy WSIs were obtained from the PANDA Challenge Dataset. We collected 10,616 core needle biopsies, each labeled with an International Society of Urological Pathology (ISUP) grade, resulting in 6 classes for multiclass classification.

\subsection{Models}
For the ultrasound downstream tasks, a linear probe was trained on each task using the predefined train/validation/test splits in Ultrabench~\cite{tupper_revisiting_2025}. Each linear probe was trained with a batch size of 32 for 300 epochs with early stopping (patience = 30) and a 0.001 learning rate. The epoch with the best validation loss was used to calculate performance on the held out test set. Given that we used the predefined splits from the benchmark, we trained each probe with five different seeds and averaged the results. 

For the pathology downstream tasks, we used attention-based multiple instance learning (ABMIL) to probe learned representations \cite{ilse2018attention}. Given the large size of WSIs and that labels are assigned at the slide or patient level, we used ABMIL to aggregate and classify the slide-level or patient-level labels. We trained an ABMIL model on each dataset for 20 epochs with a weight decay of $1\times10^{-4}$, a learning rate of $1\times10^{-3}$, and a batch size of 1 (standard practice to fit all patch embeddings from a single bag, representing one slide or patient). The best epoch was selected using validation AUROC, and this checkpoint was applied to the holdout test set. Similar to the approach used for ultrasound, we used predefined train/val/test splits and we retrained each ABMIL model with five different seeds, averaging the results to capture variation across initializations.

\subsection{Attention maps}\label{sec:sup_attn_maps}

\begin{figure}[!tbp]
  \centering
  \includegraphics[trim={0cm 1.5cm 0cm 0cm}, clip, width=\textwidth]{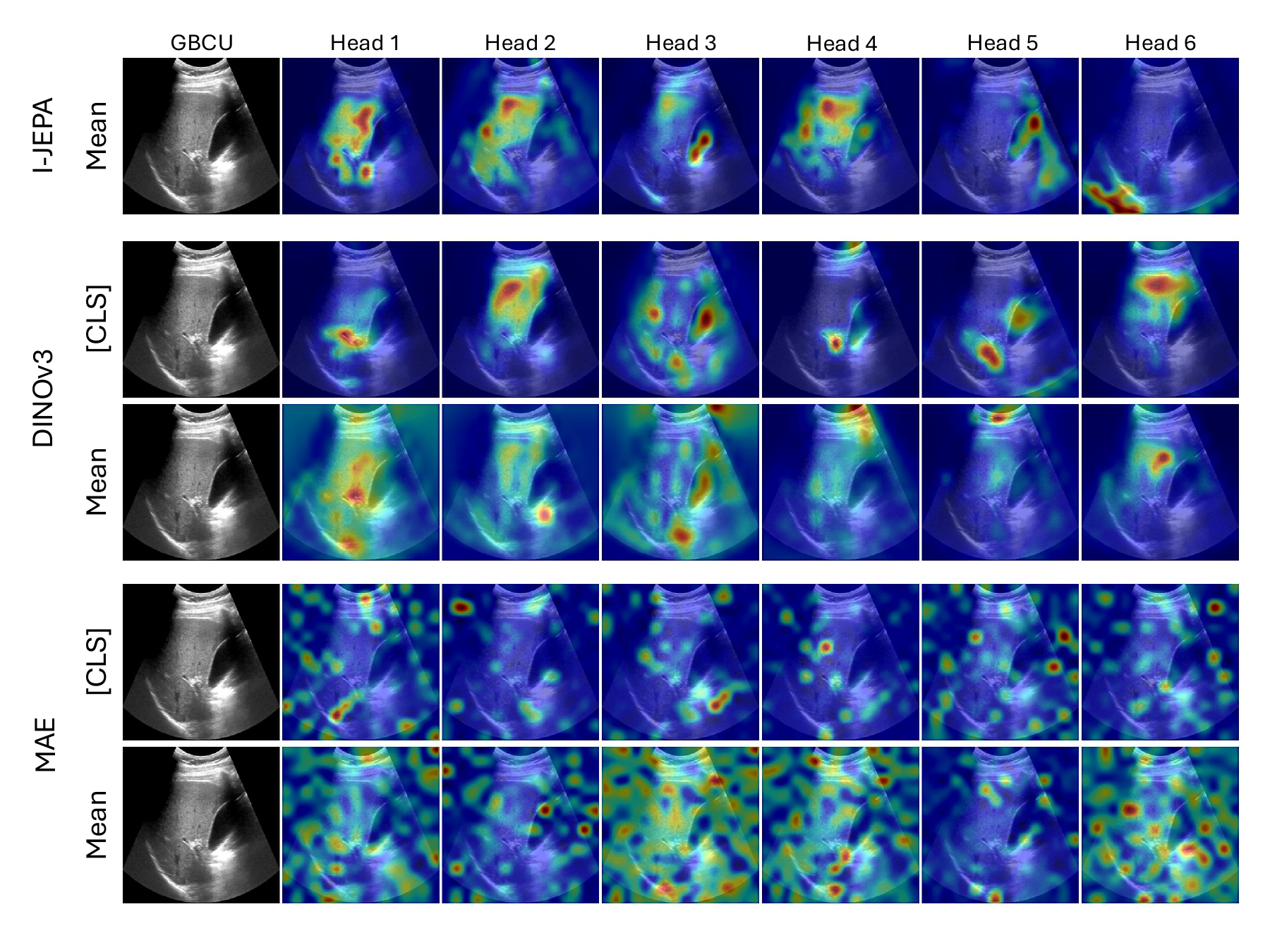}
  \caption{Example attention maps from DINOv3 and MAE comparing \texttt{[CLS]} token and mean patch attentions for an ultrasound example. I-JEPA mean patch attention map is provided as a reference.
  }
  \label{fig:sup_mean_vs_cls}
\end{figure}

\begin{figure}[!tbp]
  \centering
  \includegraphics[trim={0cm 0cm 0cm 0cm}, clip, width=\textwidth]{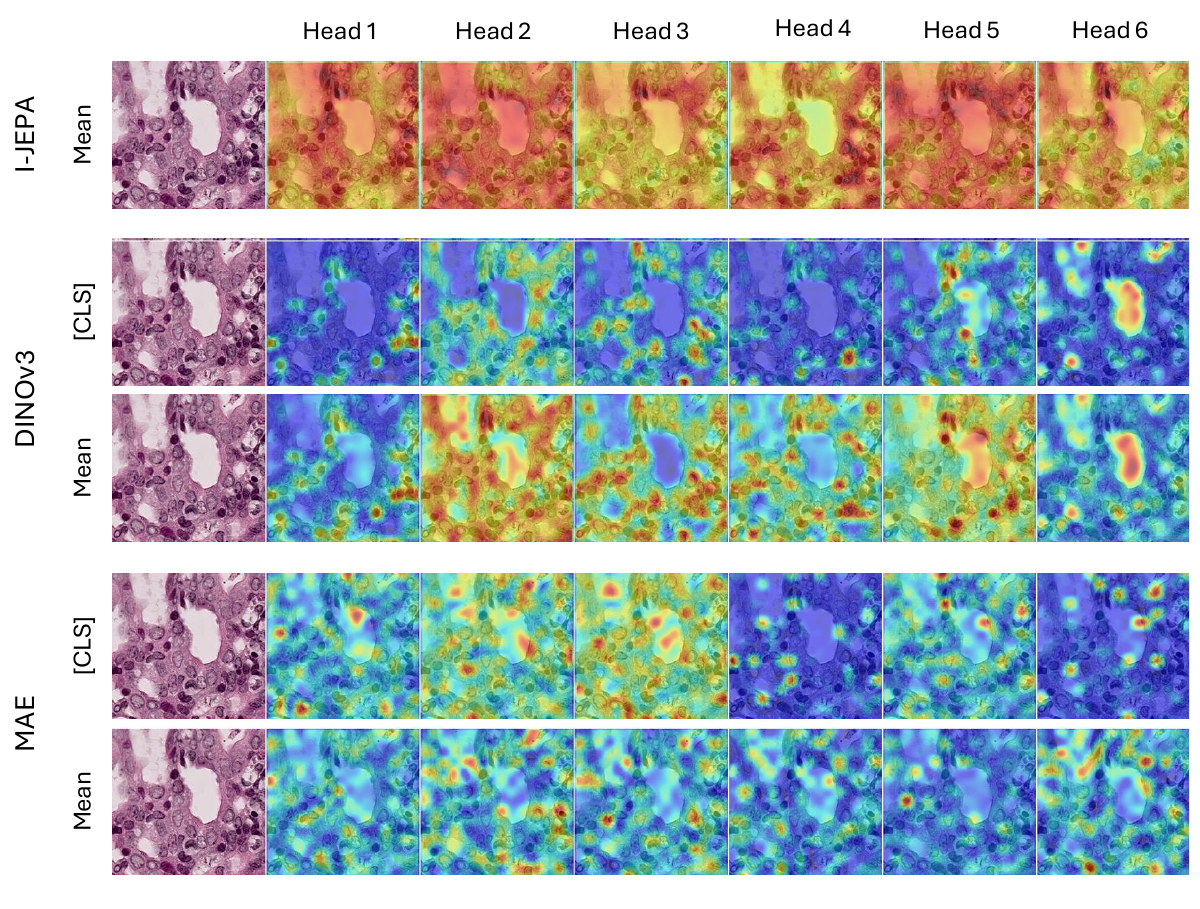}
  \caption{Example attention maps from DINOv3 and MAE comparing \texttt{[CLS]} token and mean patch attentions for a histopathology example. I-JEPA mean patch attention map is provided as a reference.
  }
  \label{fig:sup_mean_vs_cls_path}
\end{figure}

\cref{fig:sup_mean_vs_cls} and \cref{fig:sup_mean_vs_cls_path} show example mean patch attention maps for DINOv3 and MAE on example ultrasound and pathology images. Compared to when using the \texttt{[CLS]} token, the mean attention over all patches leads to less focused attention maps and activations in the non ultrasound regions and surrounding cellular area in the pathology images. We chose to use the \texttt{[CLS]} token to follow the default implementation used in the original publications to represent the models in the best light.

MAE's attention maps are dispersed with minimal clinically relevant attention. The cosine similarity maps have high similarity also dispersed across the entire image rather than in clinically relevant regions. As a result, when compared to I-JEPA, MAE is significantly worse at distinguishing semantic similarity. In identifying anatomical similarity within the liver, MAE produces significantly less semantically consistent feature clusters than I-JEPA ($p=7.15\times10^{-26}$). Across anchors within the liver, MAE is unable to learn semantic coherence of anatomical structures with generally no correlation between anchors.

\subsection{Additional Model Comparisons} 
We implemented the original I-JEPA framework; however, in the recent work LeJEPA~\cite{balestriero_lejepa_2025}, a SIGReg regularizer was introduced to stabilize training and prevent representation collapse. Given the low performance of I-JEPA in our histopathology experiments, we additionally implemented I-JEPA with the same SIGReg regularization. Despite adding this regularization, model performance did not improve across any of the downstream pathology tasks, MAE and DINOv3 still outperform I-JEPA both with and without SIGReg. MAE's strong performance over JEPA-based approaches may stem from its ability to capture finer textures through the masked reconstruction objective; even when the reconstructed details are not always task-relevant, the additional low-level focus appears to benefit downstream classification.

\section{Additional Clinical Interpretations}

\subsection{Ultrasound Examples} 

\subsubsection{Gallbladder}

\begin{figure}[!tbp]
  \centering
  \includegraphics[trim={0cm 0cm 0cm 0cm}, clip, width=\textwidth]{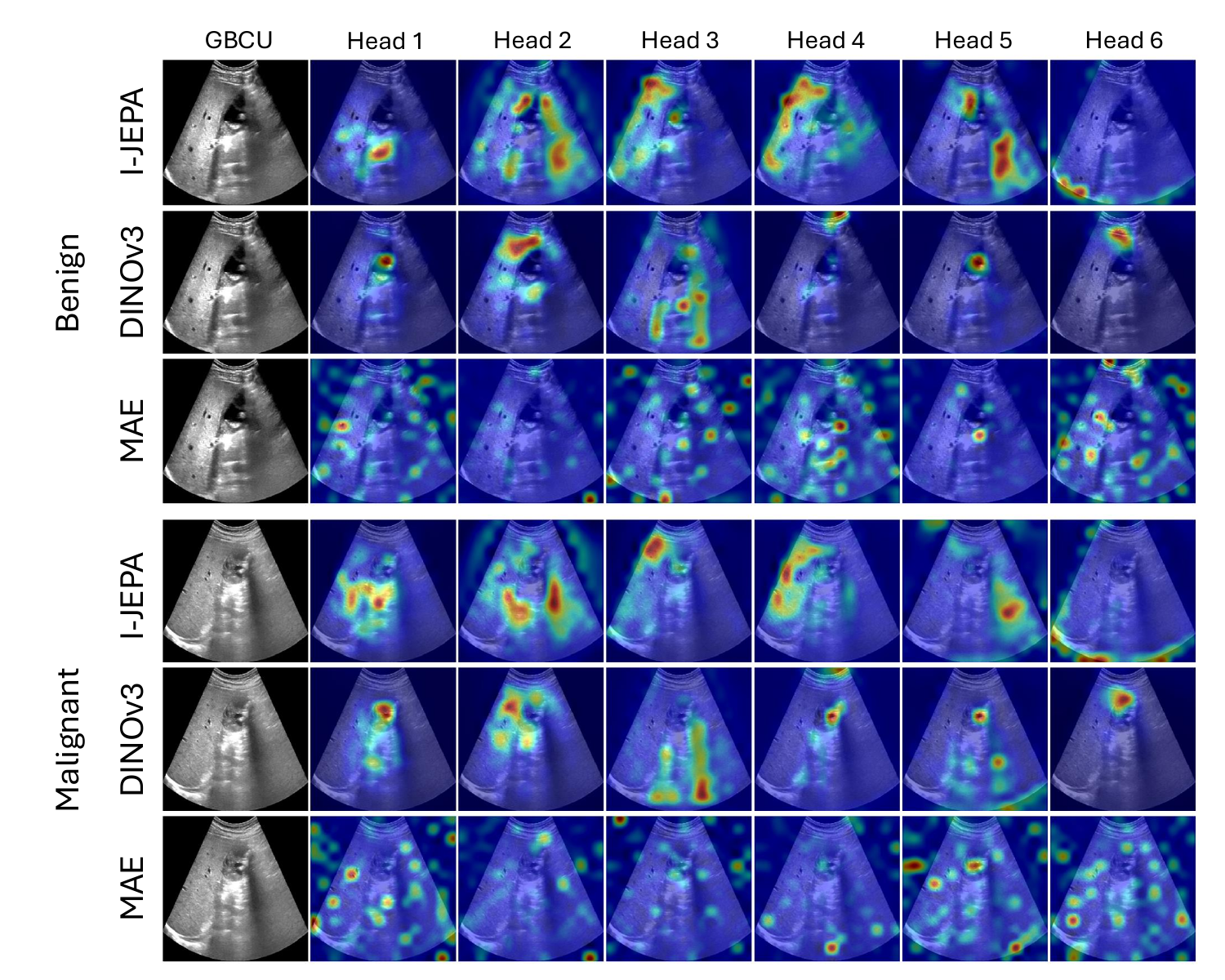}
  \caption{Attention maps from each SSL method for gallbladder ultrasounds, including both a benign and malignant example.
  }
  \label{fig:sup_gbcu}
\end{figure}

Example attention maps for both a benign and malignant gallbladder ultrasound are shown in~\cref{fig:sup_gbcu}. In the benign example, a board-certified radiologist noted that I-JEPA and DINOv3 both attend to the gallbladder and liver regions with DINOv3 attending to more localized areas and I-JEPA attending to multiple structures in a single head. DINOv3 attends to a single region at a time, for instance in the first and fifth heads the gallbladder is the focus while the second head focuses on a localized portion of the liver. I-JEPA focuses on the liver and gallbladder separately with the fourth and fifth heads, respectively. However, I-JEPA also attends to both the liver and gallbladder regions with the third head and to both the liver and non-anatomical regions with the second head. In the malignant example, DINOv3 attends to the gallbladder with the final head and the first, fourth, and fifth heads focus on the gallbladder along with some artifact. The second head focuses on a portion of the liver and additional artifact. I-JEPA primarily attends to the liver (heads 1-4), ultrasound region boundaries (head 6), and artifact (head 5). Heads 1-4 also have modest attention to the gallbladder region. Across both examples, MAE has scattered attention which does not follow clinically relevant structure.

\subsubsection{Fatty Liver}

\begin{figure}[!tbp]
  \centering
  \includegraphics[trim={0cm 0cm 0cm 0cm}, clip, width=\textwidth]{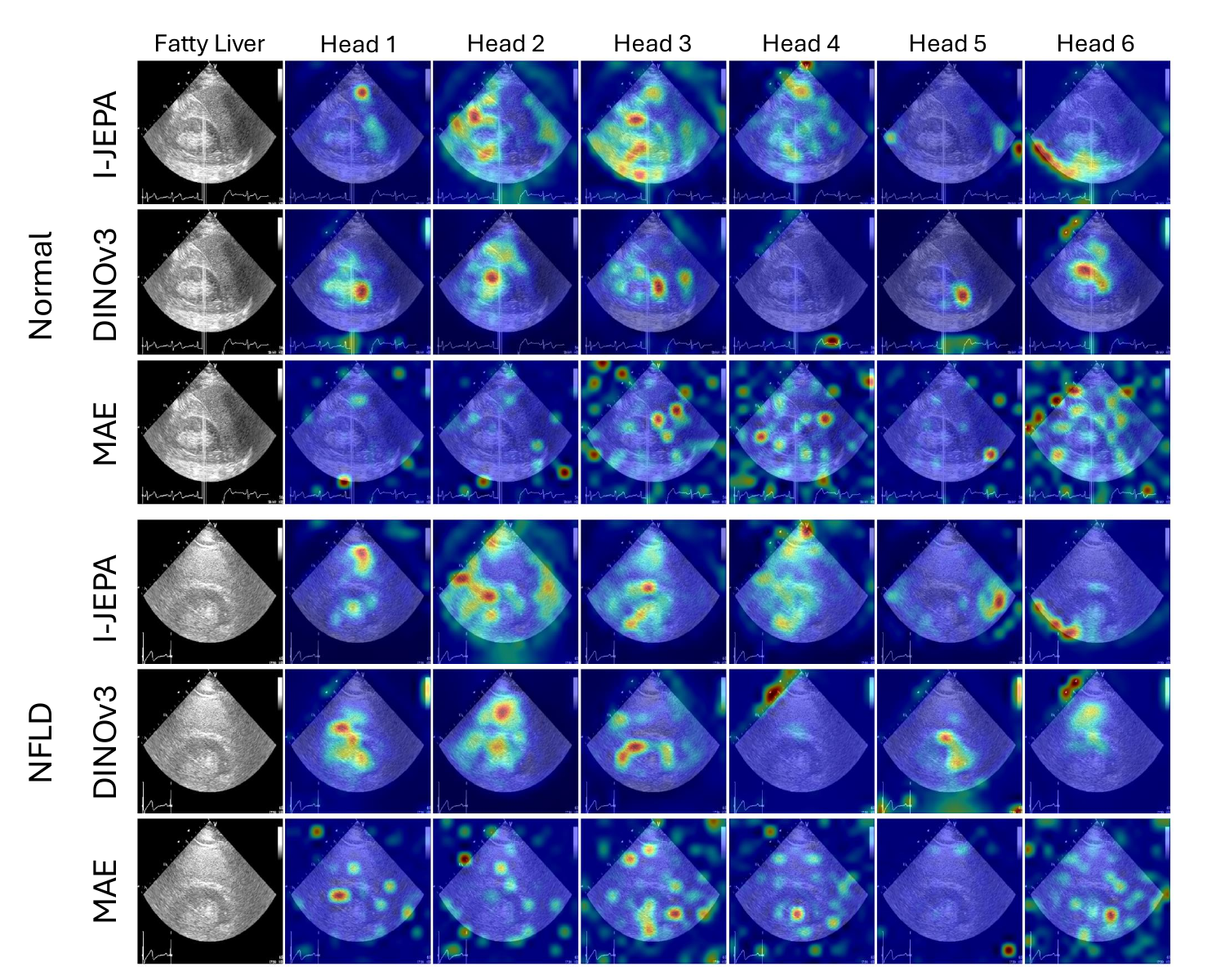}
  \caption{Attention maps from each SSL method for ultrasounds of normal liver and non-alcoholic fatty liver disease (NFLD).
  }
  \label{fig:sup_fatty_liver}
\end{figure}

\cref{fig:sup_fatty_liver} shows attention maps for ultrasounds of a normal liver and a non-alcoholic fatty liver (NFLD). MAE attention areas are consistently dispersed across the image with little correlation to the anatomical structures. However, in the NFLD ultrasound a small area of the right side kidney edge is attended to with both the second and sixth attention heads. Across ultrasounds, DINOv3 primarily attends to the kidney region with the first, third, and fifth attention heads while the second head mainly attends to the liver with modest attention to the kidney. Compared to DINOv3, I-JEPA has less localized attention as seen in the second, third, and fourth attention heads. In the second head, I-JEPA attends to the kidney and surrounding regions of the liver while the third and fourth heads attend both to the liver and kidney. In fatty liver diagnosis where the entire liver parenchyma is relevant, I-JEPA's greater attention to the liver is preferable compared to DINOv3 which primarily attends to the kidney which is irrelevant to the diagnosis.

\subsubsection{Thyroid}

\begin{figure}[!tbp]
  \centering
  \includegraphics[trim={0cm 0cm 0cm 0cm}, clip, width=\textwidth]{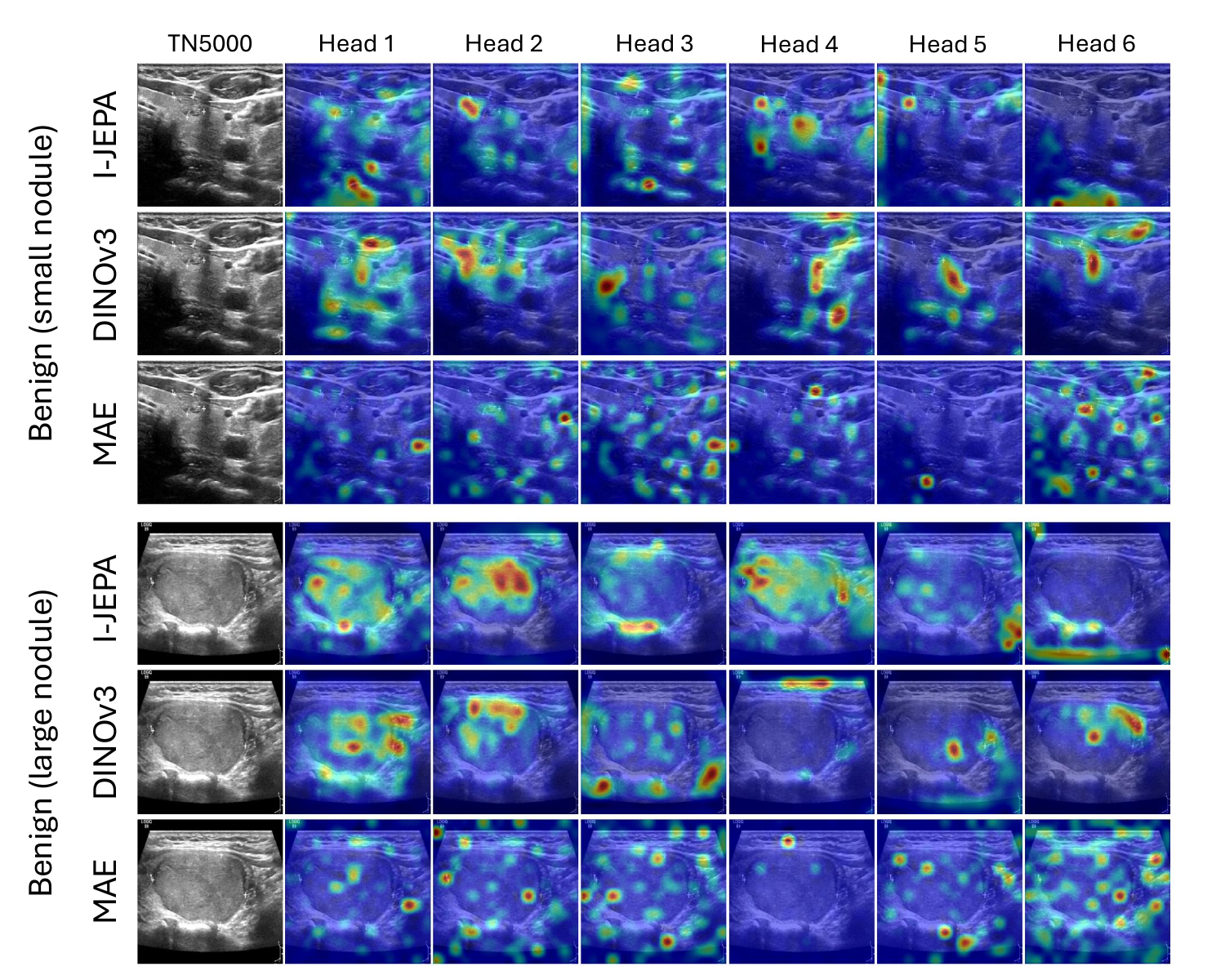}
  \caption{Attention maps from each SSL method for thyroid ultrasounds. Both ultrasounds show benign nodules but of varying sizes. Nodules are denoted by the calipers.
  }
  \label{fig:sup_thyroid}
\end{figure}

Example benign thyroid ultrasounds including both a small and large nodule are shown in~\cref{fig:sup_thyroid}. In the ultrasound of the small benign nodule, DINOv3 and I-JEPA attend to anatomical boundaries and modestly attend to small areas of the nodule. Both models attend to the boundary between the thyroid and trachea which radiologists examine when identifying nodules. MAE attends to two small areas within the nodule but the remaining attention lacks interpretable structural relevance. In the large nodule ultrasound, I-JEPA attends to larger proportions of the nodule while DINOv3 has more localized attention likely attending to textural features. Both models attend to posterior and anterior margins of the thyroid and nodule as well as various artifacts such as the edge of the ultrasound field of view. These attention patterns demonstrate that even with no fine-tuning, both DINOv3 and I-JEPA attend to clinically relevant areas for thyroid nodule detection and diagnosis.

\subsubsection{Organ differentiation}

\begin{figure}[!tbp]
  \centering
  \includegraphics[trim={0cm 0cm 0cm 0cm}, clip, width=\textwidth]{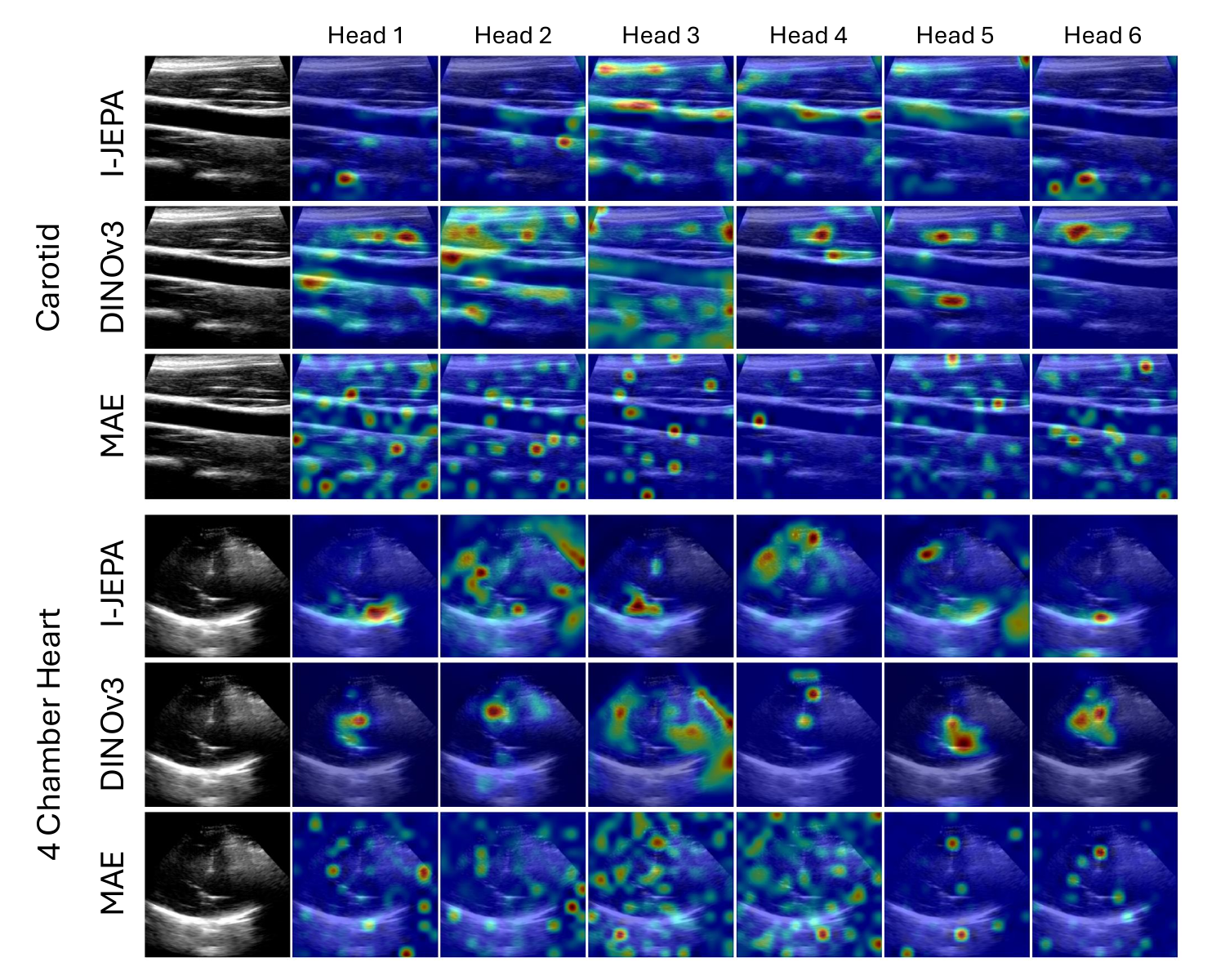}
  \caption{Attention maps from each SSL method for ultrasounds of the carotid and the four chamber view of the heart.
  }
  \label{fig:sup_region_clf}
\end{figure}

I-JEPA outperforms DINOv3 in identifying carotid and lung ultrasounds while DINOv3 outperforms I-JEPA in identifying ultrasounds of the two and four chamber views of the heart. ~\cref{fig:sup_region_clf} shows representative ultrasound images and attention maps for two of these regions: the carotid artery and a four chamber view of the heart. Across both regions, MAE attention maps do not align with clinically relevant structures.

\begin{figure}[!tbp]
  \centering
  \includegraphics[trim={0cm 0cm 0cm 0cm}, clip, width=\textwidth]{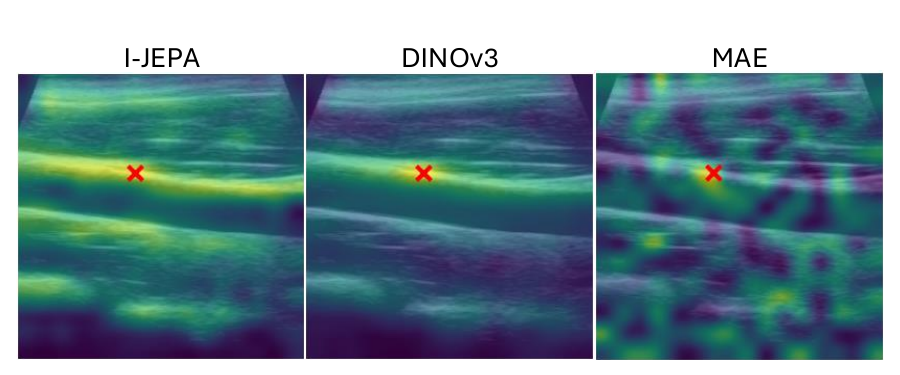}
  \caption{Carotid artery ultrasound cosine similarity map for each model. The anchor point (red X) is placed on the boundary of the carotid artery to demonstrate I-JEPA's ability to learn anatomical structures.
  }
  \label{fig:sup_cos_sim_carotid}
\end{figure}

In the carotid ultrasound attention maps, I-JEPA attends to the boundaries of the carotid artery with limited attention to surrounding tissue. I-JEPA's learned structural representations are further evident in the cosine similarity maps (\cref{fig:sup_cos_sim_carotid}), where both arterial walls cluster in representation space. In contrast, DINOv3 shows modest attention to regions of the carotid boundary and primarily focuses on surrounding tissue. In the cosine similarity map, DINOv3 identifies similarity mainly along the same border as the anchor patch, but does not strongly associate the opposite arterial wall as a similar structure. These findings suggest that I-JEPA outperforms DINOv3 on carotid region identification because it learns a more coherent structural representation of the artery. 

In the four chamber view of the heart, DINOv3 attends to the heart region while I-JEPA primarily attends to surrounding areas. This suggests that the more localized representations learned by DINOv3 are better aligned for identifying this region while broader spatial context is less informative for the task.

\subsection{Histopathology Examples} 

\begin{figure}[!tbp]
  \centering
  \includegraphics[trim={0cm 0cm 0cm 0cm}, clip, width=\textwidth]{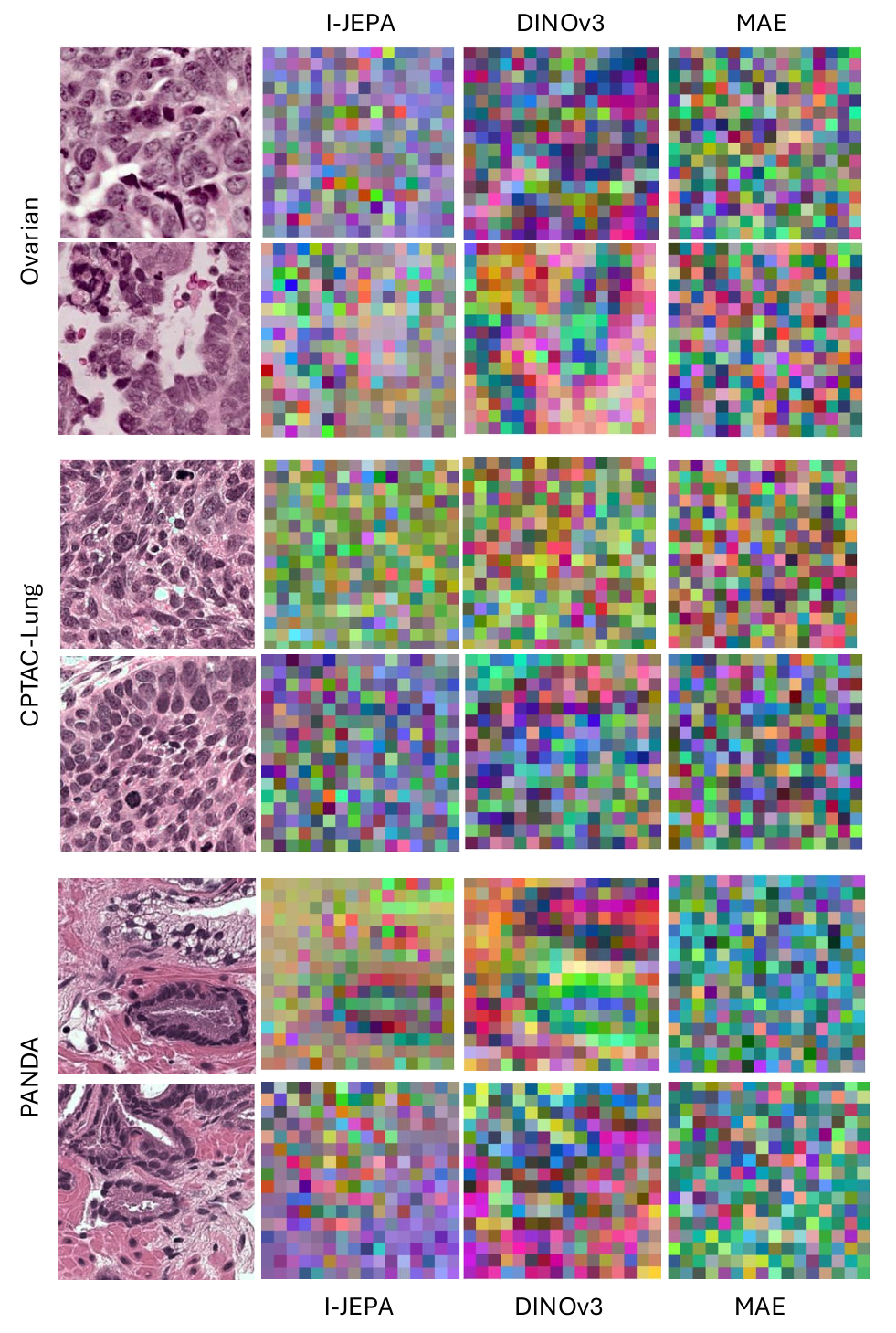}
  \caption{PCA Visualizations of two histopathology patches from three downstream examples.
  }
  \label{fig:sup_path_pca}
\end{figure}

For further qualitative assessment, we had pathologists review PCA projections of patch-level features across ovarian, lung, and prostate cancer samples (\cref{fig:sup_path_pca}). Both pathologists consistently found that MAE produced representations that appeared random or noisy, with little meaningful correspondence to the tissue. In contrast, I-JEPA and DINOv3 produced more interpretable feature maps. I-JEPA could distinguish broad structural regions, while DINOv3 demonstrated finer-grained patterns, differentiating glands and the regions within, and distinct tissue bands.

To evaluate the quality of learned representations that were not dependent on ABMIL attention weights, we also visualized attention head heatmaps and PCA projections on high-detail patches selected via Laplacian of Gaussian (LoG) filtering. We compute the LoG response of each histopathology patch and selected those with the highest values, favoring patches with rich structural content. A pathologist then reviewed the resulting visualizations across two downstream tasks, PLCO Lung and PLCO Breast. I-JEPA's PCA projections effectively categorized tissue into broad functional regions, distinguishing glands from stroma and identifying higher-level structures such as blood vessels. However, similar to the findings in the main text, I-JEPA's individual attention heads were often described as diffuse. DINOv3 demonstrated greater cellular specificity, in line with prior analysis. MAE was consistently the weakest, PCA projections appeared near-random, and attention heads lacked visible biological patterns, with only occasional and non-specific focus.

\end{document}